\documentclass[10pt,twocolumn,letterpaper]{article}

\usepackage[pagenumbers]{cvpr}      % To produce the REVIEW version
\usepackage{microtype}
\usepackage{booktabs}                                   
\usepackage{colortbl}                                    
\usepackage{tabularx}                                    
\usepackage{multirow}
\usepackage{bbm}
\newcommand{\res}[2]{#1~(#2)}                           
      
\newcommand{\dres}[2]{$#1$~($#2$)}                       
\newcommand{\fst}[1]{\textbf{#1}}                        
\newcommand{\snd}[1]{\underline{#1}}                    

\definecolor{oursbg}{RGB}{232,240,250}

\usepackage{algorithm}
\usepackage{algpseudocode}
\algrenewcommand\algorithmicrequire{\textbf{Input:}}    
\algrenewcommand\algorithmicensure{\textbf{Output:}}

\definecolor{cvprblue}{rgb}{0.21,0.49,0.74}
\usepackage[pagebackref,breaklinks,colorlinks,allcolors=cvprblue]{hyperref}

\hypersetup{
  pdftitle={HSTTA: Hierarchical Selective Test-Time Adaptation for Label-Free Cross-Subject EEG-to-Image Retrieval},
  pdfauthor={Hongzhou Chen, Yuhao Wei, Lianghua He, Yihang Liu, Zengqiang Wang}
}

\def\confName{CVPR}
\def\confYear{2027}

\title{HSTTA: Hierarchical Selective Test-Time Adaptation for Label-Free Cross-Subject EEG-to-Image Retrieval}

\author{
Hongzhou Chen\textsuperscript{1,*}\quad
Yuhao Wei\textsuperscript{1,*}\quad
Lianghua He\textsuperscript{1,2,\dag}\quad
Yihang Liu\textsuperscript{1}\quad
Zengqiang Wang\textsuperscript{1}\\
\textsuperscript{1}School of Computer Science and Technology, Tongji University\\
\textsuperscript{2}Shanghai Eye Disease Prevention and Treatment Center\\
{\tt\small \{chenhongzhou, 2353242, helianghua, 2111131, zengqiangw\}@tongji.edu.cn}
}

\begin{document}
\maketitle

\begingroup
\renewcommand\thefootnote{}
\footnotetext{\textsuperscript{*}Equal contribution.\quad \textsuperscript{\dag}Corresponding author.}
\endgroup

\begin{abstract}
Cross-subject EEG-to-image retrieval suffers from substantial performance degradation on new subjects due to subject shift.
Test-time adaptation (TTA) can mitigate this shift by adapting pretrained models with unlabeled target-subject EEG.
However, most existing TTA methods struggle to match supervision granularity to each query's prediction confidence, leaving many queries without reliable pseudo-labels.
To address this, we propose a novel framework named Hierarchical Selective Test-Time Adaptation (HSTTA).
HSTTA builds a hierarchy over gallery image embeddings, uses hierarchical selective prediction to assign each EEG query a supervision granularity matching its current prediction confidence, and constructs adaptation objectives accordingly.
We further propose Cross-Session Consistency Calibration (CSCC), an unsupervised method that calibrates the global confidence threshold to target a predefined accuracy level for hierarchical predictions.
Moreover, we introduce Reciprocal Hierarchical Backoff (RHB), which moves queries lacking reverse-matching support to coarser nodes to mitigate error reinforcement.
Experiments on two public benchmarks and ablation studies validate the effectiveness of HSTTA and its key components.
On THINGS-EEG2, HSTTA outperforms the strongest EEG-specific TTA baseline by 17.9 (13.4) and 8.9 (3.7) percentage points in Top-1/Top-5 accuracy, respectively, under cosine (CSLS) ranking.
Further analyses show that HSTTA yields more accurate pseudo-labels during adaptation, supporting more reliable model updates.
\end{abstract}    
\section{Introduction}
\label{sec:intro}

Visual brain decoding aims to infer stimulus content from the corresponding neural responses~\cite{kay2008identifying}.
Electroencephalography (EEG) is increasingly used in visual brain decoding, including classification, image retrieval~\cite{ICLR2026_c5506f15}, and reconstruction~\cite{yu2026d}, owing to its portability, low cost, and high temporal resolution~\cite{mushtaq2024one}.
In EEG-to-image retrieval, mainstream methods align EEG and image representations through contrastive learning and rank candidate images by similarity.
On the widely used THINGS-EEG2~\cite{gifford2022large} benchmark, advanced methods have achieved high intra-subject Top-1 accuracy (over 90\% in 200-way retrieval~\cite{jiang2026subject}).
However, a substantial performance gap persists in cross-subject settings (Top-1 below 20\% for most representative baselines~\cite{song2024decoding, chen2024visual, wu2025bridging}).
This gap limits deployment, making improved cross-subject decoding an important step toward practical EEG-based visual brain–computer interfaces.

\begin{figure}[t]
  \centering
  \includegraphics[width=\linewidth]{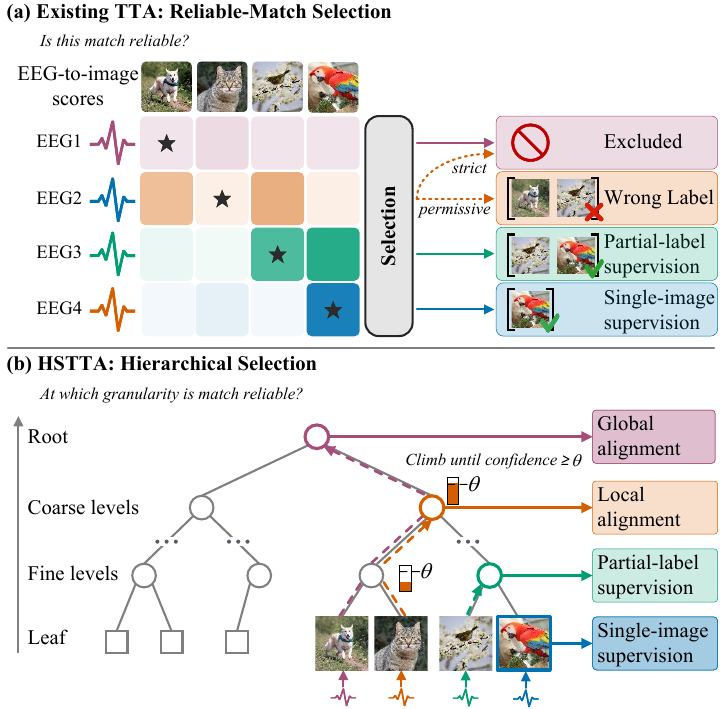}
  \caption{
    Supervision granularity for EEG queries with varying confidence.
    (a) Existing TTA methods typically assign selected queries single-image or partial-label supervision, excluding the rest.
    (b) HSTTA uses the gallery hierarchy to select an appropriate supervision granularity based on each query's current confidence.
  }
  \label{fig:motivation}
\end{figure}

The cross-subject gap stems mainly from subject shift, where individual differences in visual cortex~\cite{lipman2025invariant}, head anatomy, and electrode placement make the same image evoke different EEG responses across subjects.
A common approach is to fine-tune a pretrained model using a few labeled EEG–image pairs from the new subject~\cite{han2026cross}, but this requires extra calibration recordings, raising deployment costs.
Test-time adaptation (TTA) instead adapts a pretrained model using unlabeled EEG from the new subject, offering a way to handle subject shift without paired labels.
Existing TTA methods adapt models through training-free feature transformations or gradient-based parameter updates~\cite{liang2025comprehensive}.
Training-free methods fix the parameters and align test feature statistics~\cite{huang2026sattc} or estimate transformations of features or similarity scores from pseudo-labels~\cite{iwasawa2021test, karmanov2024efficient, cui2026score}. 
Gradient-based methods typically update a subset of parameters by minimizing entropy~\cite{wang2021tent, li2025tcr, lafon2025cliptta} or cross-entropy with pseudo-labels~\cite{hu2025selective}.
Therefore, pseudo-label reliability largely determines adaptation performance, making reliable selection central to TTA research.

In cross-subject retrieval, single-image matches for new subjects are often unreliable, so directly applying entropy minimization or pseudo-labeling may induce confirmation bias~\cite{zhang2026uatta}, reinforcing errors and degrading retrieval performance.
To mitigate this, existing methods typically select reliable matches using adaptive entropy thresholds~\cite{li2025tcr}, bidirectional consistency~\cite{cui2026score, zhang2026uatta}, or confidence margins~\cite{hu2025selective}.
As Fig.~\ref{fig:motivation}(a) shows, selected queries receive single-image or partial-label supervision over candidate images~\cite{hu2025selective}, while less reliable ones are downweighted or excluded.
However, many EEG queries lack sufficient confidence to support reliable single-image matches, yet the model may still associate them with a group of similar images.
For these queries, permissive filtering may introduce incorrect single-image supervision, whereas strict filtering may exclude them, leaving coarse-grained matching information unused.
Thus, a key limitation of existing methods is their limited supervision granularity, which prevents them from fully adapting the supervision level to each query's confidence.

To address this limitation, inspired by hierarchical selective classification (HSC)~\cite{goren2024hierarchical}, we propose Hierarchical Selective Test-Time Adaptation (HSTTA), a novel framework that builds a hierarchical tree over gallery image embeddings and assigns each EEG query a supervision granularity appropriate to its confidence.
As shown in Fig.~\ref{fig:motivation}(b), the model may be uncertain whether a query matches a cat image but still confidently associate it with a group of animal images, enabling group-level adaptation.
Specifically, we adopt HSC's climbing rule, tracing parent links from the best-matching leaf to the first node meeting a global confidence threshold.
Since calibrating HSC's threshold requires labels unavailable in our TTA setting, we propose Cross-Session Consistency Calibration (CSCC), which calibrates the threshold without labels using cross-session prediction consistency of repeated EEG responses to the same stimulus.
% Moreover, we propose Reciprocal Hierarchical Backoff (RHB), which backs off queries lacking reverse-matching support to coarser nodes, mitigating the reinforcement of incorrect pseudo-labels during offline TTA.
To mitigate the reinforcement of incorrect pseudo-labels during offline TTA, we further propose Reciprocal Hierarchical Backoff (RHB), which backs off queries lacking reverse-matching support to coarser nodes.
The final selected nodes and their paths define the adaptation objectives.

In summary, our main contributions are as follows:
\begin{itemize}
    \item To the best of our knowledge, we are the first to apply hierarchical selective prediction to TTA. We propose HSTTA, which assigns each unlabeled EEG query a supervision granularity appropriate to its prediction confidence and constructs adaptation objectives accordingly.
    \item  We propose CSCC and RHB to support reliable hierarchical selective prediction without labels.
    CSCC calibrates the new subject's global confidence threshold, while RHB mitigates error reinforcement during adaptation.
\item HSTTA substantially improves pretrained SAMGA's~\cite{jiang2026subject} cross-subject retrieval, achieving the highest Top-1 accuracy among evaluated methods on THINGS-EEG2~\cite{gifford2022large} and Alljoined-1.6M~\cite{xu2025alljoined} in standard retrieval. Experiments with disjoint galleries and other encoders demonstrate its applicability. Adaptation analyses show higher pseudo-label selection accuracy than baselines.
\end{itemize}
\begin{figure*}[ht]
  \centering
  \includegraphics[width=\textwidth]{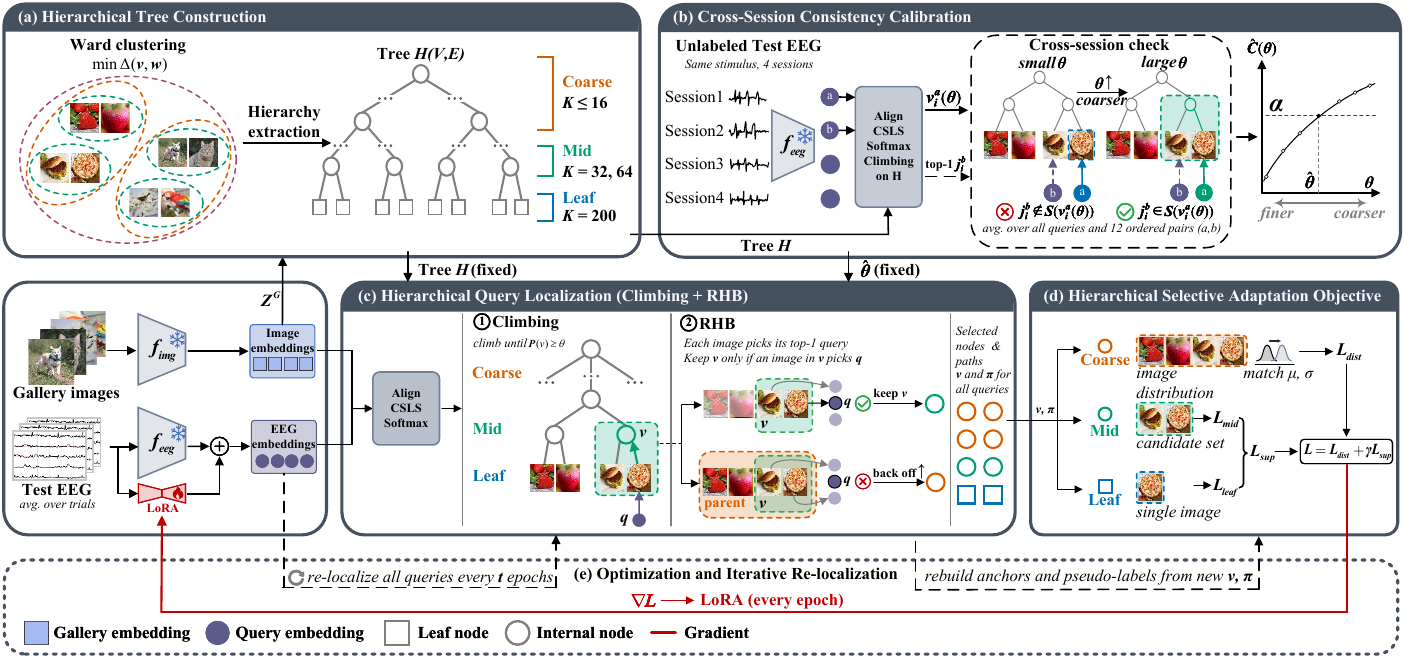}
    \caption{Overview of HSTTA. (a)(b)~Ward tree $H$ construction and threshold $\hat{\theta}$ calibration before adaptation. (c)~Localization of each query to its finest reliable level. (d)~Level-wise supervision. (e)~Per-epoch LoRA updates with re-localization every $t$ epochs.}
  \label{fig:framework}
\end{figure*}

\section{Related Work}
\label{sec:related_work}

\subsection{Cross-Subject Brain-to-Image Retrieval}
Cross-subject brain-to-image retrieval transfers models trained on source subjects to unseen subjects~\cite{wang2024mindbridge, huang2026sattc}, but inter-subject neural variability degrades accuracy~\cite{ICLR2026_c5506f15, jiang2026subject}.
Some studies introduce subject-specific calibration modules that require labeled target-subject data before deployment~\cite{gong2025mindtuner, li2026duala, liu2026mindadapter, xu2026pda_lora}, while others learn subject-invariant representations~\cite{wang2025zebra, jiang2026subject}.
Recent studies explore label-free test-time calibration to improve cross-subject EEG-to-image retrieval.
SATTC~\cite{huang2026sattc} recalibrates similarity scores through whitening, adaptive CSLS~\cite{lample2018word}, and mutual-rank structural priors.
SCORE~\cite{cui2026score} recovers target-subject coordinates through orthogonal Procrustes alignment using mutual nearest-neighbor pairs.
Unlike SCORE's shared global transformation, HSTTA selects pseudo-label granularity according to each query's prediction confidence and adapts the pretrained EEG encoder through gradient updates.

\subsection{Hierarchical Selective Prediction}
Selective prediction allows models to abstain when confidence is insufficient~\cite{el2010foundations}, whereas hierarchical selective prediction backs off to a coarser node of a class hierarchy when fine-grained predictions are uncertain~\cite{deng2012hedging, wu2020solving}.
In image classification, DARTS~\cite{deng2012hedging} predicts the node with the highest transformed expected reward on a WordNet-derived~\cite{miller1995wordnet} ImageNet~\cite{deng2009imagenet} hierarchy, tuning its trade-off parameter $\lambda$ on validation set to meet a target accuracy.
Valmadre~\cite{valmadre2022hierarchical} predicts the most informative node whose confidence exceeds a threshold and evaluates classifiers across all thresholds.
HSC~\cite{goren2024hierarchical} climbs from the most probable leaf to the first node on its path to the root whose confidence reaches a global threshold, calibrated on labeled data to meet a target accuracy.
However, these methods set $\lambda$ or the confidence threshold using labeled data from the test distribution, unavailable for an unseen subject.
CSCC instead calibrates the target subject's threshold without labels using cross-session prediction consistency.

\subsection{Test-Time Adaptation}
Test-time adaptation adapts pretrained models to distribution shifts using unlabeled data.
Among training-free methods, T3A~\cite{iwasawa2021test} adjusts the classifier using class-wise pseudo-prototypes, while TDA~\cite{karmanov2024efficient} refines predictions through positive and negative pseudo-label caches.
Among gradient-based methods, TENT~\cite{wang2021tent} updates normalization-layer affine parameters through entropy minimization.
For cross-modal tasks, TCR~\cite{li2025tcr} addresses query shift through prediction refinement and representation regularization, CLIPTTA~\cite{lafon2025cliptta} uses a soft contrastive objective, and UATTA~\cite{zhang2026uatta} guides adaptation using bidirectional retrieval disagreement.
PASLE~\cite{hu2025selective}, our closest baseline, assigns single pseudo-labels to confident samples and candidate label sets to uncertain ones, progressively refining them during adaptation.
However, its candidate sets are flat, margin-based sets that ignore the structure among gallery images, and in offline adaptation, where the same queries are revisited over many epochs, its progressive refinement can reinforce early pseudo-label errors.
HSTTA instead selects each query's supervision granularity on a hierarchy built from gallery embeddings via HSC~\cite{goren2024hierarchical}, while RHB backs off queries that fail reverse matching checks to coarser ancestor nodes.

\section{Hierarchical Selective Test-Time Adaptation}

\subsection{Problem Setup and Overview}
Given a model pretrained on source subjects, we aim to retrieve the matching gallery image for each unlabeled EEG query from a target subject.
Let $\mathcal Q=\{q_i\}_{i=1}^{N^Q}$ and $\mathcal G=\{g_j\}_{j=1}^{N^G}$ denote the query set and image gallery, respectively.
Each query $q_i=\{x_i^r\}_{r=1}^{R_i}$ contains $R_i$ repeated EEG trials elicited by the same stimulus across sessions.
During TTA, stimulus grouping is known, but EEG--image correspondences are unavailable.
The EEG encoder $f_{\mathrm{eeg}}$ encodes trial-averaged EEG, while the frozen image encoder $f_{\mathrm{img}}$ produces gallery embeddings:
\begin{equation}
    \mathbf z_i^Q
    =
    f_{\mathrm{eeg}}\!\left(
        \frac{1}{R_i}\sum_{r=1}^{R_i}x_i^r
    \right),
    \quad
    \mathbf z_j^G=f_{\mathrm{img}}(g_j),
\end{equation}
where $\mathbf z_i^Q,\mathbf z_j^G\in\mathbb R^d$.
During TTA, we freeze the image encoder and adapt the EEG encoder via LoRA~\cite{hu2022lora} together with its normalization-layer affine parameters, using all queries and gallery images, and then rank gallery images by the cosine similarity between $\mathbf{z}_i^Q$ and $\mathbf{z}_j^G$.

Fig.~\ref{fig:framework} illustrates HSTTA's overall framework.
We first build a hierarchical tree over the gallery image embeddings (Sec.~\ref{sec:wardtree_construct}) and localize each query with HSC's Climbing rule under a global confidence threshold $\theta$ (Sec.~\ref{sec:hierarchical_prediction}), which is calibrated by CSCC without labels and kept fixed (Sec.~\ref{sec:cscc}).
The adaptation objective of each query is determined by the level of its node (Sec.~\ref{sec:objective}).
During TTA, we re-localize all queries every $t$ epochs, back off queries lacking reverse-matching support via RHB, and continue optimizing the EEG encoder with the updated supervision (Sec.~\ref{sec:relocalization}).

\subsection{Hierarchical Tree Construction from Gallery Image Embeddings}
\label{sec:wardtree_construct}
HSC uses a predefined class hierarchy, such as WordNet for ImageNet, with specific classes at leaves and broader categories at internal nodes.
In EEG-to-image retrieval, each prediction targets an individual gallery image.
We therefore group gallery images into nested sets using embedding distances, without an external taxonomy (Fig.~\ref{fig:framework}(a)).

\textbf{Ward clustering.}
We apply Euclidean Ward agglomerative clustering to gallery embeddings $\{\mathbf z_j^G\}_{j=1}^{N^G}$~\cite{ward1963hierarchical}.
Starting from $N^G$ singleton clusters, we repeatedly merge the pair $(v,w)$ minimizing the increase in the sum of squared Euclidean distances to cluster means:
\begin{equation}
    \Delta(v,w)
    =
    \frac{m_vm_w}{m_v+m_w}
    \left\|\boldsymbol\mu_v^G-\boldsymbol\mu_w^G\right\|_2^2,
\end{equation}
where $m_v,m_w$ are the cluster sizes and $\boldsymbol\mu_v^G,\boldsymbol\mu_w^G$ their mean embeddings.
After $N^G-1$ merges, we obtain the full clustering tree with individual gallery images as leaves and the entire gallery as its root.

\textbf{Hierarchy extraction.}
We retain groupings with $K\in\{1,2,4,8,16,32,64,N^G\}$ clusters from the Ward merge sequence. 
Coarse clusters are unions of finer clusters, so these groupings are nested. 
Clusters with identical members across levels are merged into one node, assigned to the finest level at which they appear.

The extracted tree $H=(V,E)$ has node set $V$ and directed edges $E$ from parents to children.
Let $\mathcal S(v)$ denote the gallery image indices in node $v$, with $m_v=|\mathcal S(v)|$.
Each non-root node's parent is the retained node with the smallest member set strictly containing $\mathcal S(v)$.
The root $v_{\mathrm{root}}$ contains all gallery images, and each leaf in $\mathcal Y\subset V$ contains one image.
The tree remains fixed during adaptation.

\subsection{Hierarchical Prediction with Climbing}
\label{sec:hierarchical_prediction}
We estimate query--gallery matching probabilities and apply Climbing on tree $H$ to assign each query a pseudo-label at the corresponding granularity (Fig.~\ref{fig:framework}(c)).

\textbf{Query--gallery matching probabilities.}
To reduce global statistical discrepancies between query and gallery embeddings, we apply non-learned global statistical alignment:
\begin{equation}
\label{eq:globalalign}
    \widehat{\mathbf z}_i^Q
    =
    \frac{\mathbf z_i^Q-\boldsymbol\mu^Q}
         {\boldsymbol\sigma^Q}
    \boldsymbol\sigma^G
    +\boldsymbol\mu^G,
\end{equation}
where $(\boldsymbol\mu^Q,\boldsymbol\sigma^Q)$ and $(\boldsymbol\mu^G,\boldsymbol\sigma^G)$ are the per-dimension means and standard deviations of all query embeddings and all gallery embeddings, respectively.

To mitigate matching bias from hubness, we adjust cosine similarities using CSLS~\cite{lample2018word}:
\begin{equation}
    c_{ij}=\cos(\widehat{\mathbf z}_i^Q,\mathbf z_j^G),
    \quad
    s_{ij}=2c_{ij}-\bar c_i^Q-\bar c_j^G,
\end{equation}
where $\bar c_i^Q$ and $\bar c_j^G$ are the means of the top-$k$
similarities in row $i$ and column $j$, respectively,
with $k=10$~\cite{cui2026score}.
A softmax with temperature $\tau$ yields matching probabilities:
\begin{equation}
\label{eq:prob}
    p_{ij}
    =
    \frac{\exp(s_{ij}/\tau)}
         {\sum_{\ell=1}^{N^G}\exp(s_{i\ell}/\tau)}.
\end{equation}

\textbf{Query semantic neighborhood localization.}
We apply HSC's Climbing rule from the most probable leaf, following parent links until the confidence threshold is met.
A hierarchical prediction is correct if the selected node contains the query's true match.
For query $q_i$, its confidence at node $v$ is the total matching probability of its member images:
\begin{equation}
\label{eq:nodeprob}
    P_i(v)=\sum_{j\in\mathcal S(v)}p_{ij}.
\end{equation}
Given a global confidence threshold $\theta\in[0,1]$, HSC's Climbing rule starts at the most probable leaf $v_i=\operatorname*{arg\,max}_{v\in\mathcal Y}P_i(v)$ and follows parent links until $P_i(v_i)\ge\theta$.
Confidence is nondecreasing toward the root, where it equals $1$, ensuring termination at the most specific node on this path satisfying the threshold.

The selected node $v_i$ defines query $q_i$'s semantic neighborhood $\{g_j:j\in\mathcal S(v_i)\}$.
A leaf provides a single-image pseudo-label, while an internal node provides a coarser candidate set.
Tracing parent links from $v_i$ to the root yields $\pi_i=(v_i,\ldots,v_{\mathrm{root}})$, representing nested semantic neighborhoods.
We use the node and its path to construct the adaptation objective.

\subsection{Cross-Session Consistency Calibration}
\label{sec:cscc}
HSC calibrates its confidence threshold using labeled data.
However, target-subject EEG--image pairing labels are unavailable, and source-calibrated thresholds may not transfer reliably under subject shift.
We therefore propose Cross-Session Consistency Calibration (CSCC) to select $\theta$ without labels by comparing predictions of repeated EEG responses to the same stimulus across sessions (Fig.~\ref{fig:framework}(b)).

Let $A\ge2$ denote the number of recording sessions and $R_i^a$ the trial count for $q_i$ in session $a$, with $\sum_{a=1}^{A}R_i^a=R_i$.
We average each query's trials per session and encode them with the pre-adaptation $f_{\mathrm{eeg}}$, obtaining $\mathbf z_i^{Q,a}$.
For each session, we apply statistical alignment, CSLS, and softmax from Sec.~\ref{sec:hierarchical_prediction} to all its query embeddings and the fixed gallery, yielding matching probabilities $p_{ij}^{a}$.
Climbing on $H$ yields node $v_i^{a}(\theta)$ for each candidate threshold $\theta$.

The top-1 gallery image index for $q_i$ in session $b$ is
\begin{equation}
    j_i^{b}=\operatorname*{arg\,max}_{j}p_{ij}^{b}.
\end{equation}
For each ordered session pair $(a,b)$ with $a\ne b$, we count $q_i$ as consistent if session $a$'s semantic neighborhood contains session $b$'s top-1 prediction, i.e., $j_i^b\in\mathcal S(v_i^a(\theta))$.
Averaging over all queries and $A(A-1)$ ordered pairs yields the cross-session consistency rate:
\begin{equation}
    \widehat C(\theta)
    =
    \frac{1}{N^Q A(A-1)}
    \sum_{i=1}^{N^Q}
    \sum_{a=1}^{A}
    \sum_{\substack{b=1\\b\ne a}}^{A}
    \mathbbm{1}\!\left[
    j_i^{b}\in\mathcal S\!\left(v_i^{a}(\theta)\right)
    \right],
\end{equation}
where $\mathbbm{1}[\cdot]$ denotes the indicator function.
This rate reflects cross-session stability of predicted semantic neighborhoods and guides label-free calibration of $\theta$.

From a threshold grid $\Theta\subseteq[0,1]$, we select the smallest threshold reaching a predefined consistency level $\alpha$:
\begin{equation}
    \widehat\theta
    =
    \min\left\{
    \theta\in\Theta:
    \widehat C(\theta)\ge\alpha
    \right\}.
\end{equation}
We perform CSCC once before adaptation and use $\widehat\theta$ for Climbing on $\mathbf z_i^Q$, encoded from the average of all $R_i$ trials.
The threshold remains fixed throughout adaptation and query re-localization.
Pseudocode for CSCC is provided in the supplementary material.

\subsection{Hierarchical Selective Adaptation Objective}
\label{sec:objective}

We construct adaptation objectives from the final node $v_i$ and path $\pi_i$ after RHB (Sec.~\ref{sec:relocalization}).
Nodes at $K\in\{1,2,4,8,16\}$, $K\in\{32,64\}$, and $K=N^G$ are coarse, mid-level, and leaf nodes, respectively.
All queries receive distribution alignment; mid-level and leaf queries additionally receive candidate-set and single-image pseudo-label supervision, respectively (Fig.~\ref{fig:framework}(d)).

\textbf{Distribution alignment.}
At anchor $v$, we align the embedding distributions of queries with $v\in\pi_i$ and gallery images in $v$, where $n_v=|\{i:v\in\pi_i\}|$.
The anchor set $V_{\mathrm{anc}}$ contains the root and non-root nodes satisfying $5\le n_v\le m_v$; the root aligns all queries with the full gallery.
We remove global scale differences through shared scalar rescaling:
\begin{equation}
    \widetilde{\mathbf z}_i^Q
    =
    \frac{\|\boldsymbol\sigma^G\|_2}
         {\|\boldsymbol\sigma^Q\|_2}\mathbf z_i^Q.
\end{equation}
At anchor $v$, let $(\widetilde{\boldsymbol\mu}_v^Q,\widetilde{\boldsymbol\sigma}_v^Q)$ and $(\boldsymbol\mu_v^G,\boldsymbol\sigma_v^G)$ denote the per-dimension means and standard deviations of the rescaled query and gallery embeddings.
To account for sampling variability, we repeatedly draw $n_v$ images without replacement from each non-root anchor's gallery subset, and set $\delta_{\mu,v}$ and $\delta_{\sigma,v}$ to the 90th percentiles of the Euclidean distances between subsample and full-subset means and standard deviations, respectively; both are zero at the root.
The distribution loss $\mathcal L_{\mathrm{dist}}=\mathcal L_{\mathrm{dist}}^{\mathrm{glob}}+\mathcal L_{\mathrm{dist}}^{\mathrm{loc}}$ sums the root (global) and non-root (local) anchor terms:
\begin{equation}
\label{eq:dist_align}
\begin{aligned}
\mathcal L_{\mathrm{dist}}
=
\sum_{v\in V_{\mathrm{anc}}}\frac{n_v}{N^Q}
\Big(
&\big[
\|\widetilde{\boldsymbol\mu}_v^Q-\boldsymbol\mu_v^G\|_2
-\delta_{\mu,v}
\big]_+^2\\
+{}&
\big[
\|\widetilde{\boldsymbol\sigma}_v^Q-\boldsymbol\sigma_v^G\|_2
-\delta_{\sigma,v}
\big]_+^2
\Big),
\end{aligned}
\end{equation}
where $[x]_+=\max(x,0)$ and $n_v/N^Q$ weights each anchor by its query count.
For coarse-node queries, $\mathcal L_{\mathrm{dist}}$ provides distribution-level supervision when finer-grained pseudo-labels are unavailable.

\textbf{Partial-label supervision.}
Mid-level nodes narrow the candidate set relative to coarse ancestors but do not identify a single image.
We use node members as candidate labels and increase their total matching probability with the classifier-consistent (CC) loss~\cite{feng2020provably, hu2025selective}.
Let $\mathcal I_{\mathrm{mid}}$ index queries at mid-level nodes.
We compute $s_{ij}^{\mathrm{tr}}=\cos(\mathbf z_i^Q,\mathbf z_j^G)/\tau$ from current embeddings, without statistical alignment or CSLS.
For $i\in\mathcal I_{\mathrm{mid}}$, we minimize:
\begin{equation}
    \mathcal L_{\mathrm{mid}}(q_i)
    =
    -\log
    \frac{\sum_{j\in\mathcal S(v_i)}\exp(s_{ij}^{\mathrm{tr}})}
         {\sum_{\ell=1}^{N^G}\exp(s_{i\ell}^{\mathrm{tr}})}.
\end{equation}

\textbf{Single-image pseudo-label supervision.}
When a query's confidence supports a leaf prediction, we use the corresponding image as a pseudo-label for EEG--image matching.
Let $\mathcal I_{\mathrm{leaf}}$ index queries at leaves.
For $i\in\mathcal I_{\mathrm{leaf}}$, let $\mathcal S(v_i)=\{j_i\}$ and compute cross-entropy over all gallery images:
\begin{equation}
    \mathcal L_{\mathrm{leaf}}(q_i)
    =
    -\log
    \frac{\exp(s_{ij_i}^{\mathrm{tr}})}
         {\sum_{\ell=1}^{N^G}\exp(s_{i\ell}^{\mathrm{tr}})}.
\end{equation}

\textbf{Overall adaptation objective.}
We average losses over queries localized to mid-level and leaf nodes:
\begin{equation}
\mathcal L_{\mathrm{sup}}
=
\frac{
\sum_{i\in\mathcal I_{\mathrm{mid}}}\mathcal L_{\mathrm{mid}}(q_i)
+
\sum_{i\in\mathcal I_{\mathrm{leaf}}}\mathcal L_{\mathrm{leaf}}(q_i)
}{
|\mathcal I_{\mathrm{mid}}|+|\mathcal I_{\mathrm{leaf}}|
}.
\end{equation}
We set $\mathcal L_{\mathrm{sup}}=0$ when both sets are empty.
The overall objective is: 
\begin{equation}
\mathcal L
=
\mathcal L_{\mathrm{dist}}
+
\gamma\,\mathcal L_{\mathrm{sup}},
\end{equation}
where $\gamma$ is set from the initial gradient norms of both losses using a preset ratio and stays fixed during TTA.

\begin{table*}[ht]
\centering
\caption{Entries are accuracy (\%) under \textit{cosine (CSLS)} ranking.
The second and third blocks list training-free and learnable TTA methods, respectively.
GA adapts with $\mathcal{L}_{\mathrm{dist}}^{\mathrm{glob}}$ only.
$p$-values in the text are from two-sided paired $t$-tests over held-out subjects.}
\label{tab:main}
\footnotesize
\setlength{\tabcolsep}{2pt} 
\renewcommand{\arraystretch}{1.05}
    \begin{tabularx}{\linewidth}{@{}l*{8}{>{\centering\arraybackslash}X}@{}}
    \toprule
    & \multicolumn{4}{c}{THINGS-EEG2} & \multicolumn{4}{c}{Alljoined-1.6M} \\
    \cmidrule(lr){2-5}\cmidrule(l){6-9}
    & \multicolumn{2}{c}{Standard (200-way)} & \multicolumn{2}{c}{Disjoint gallery (100-way)}
    & \multicolumn{2}{c}{Standard (200-way)} & \multicolumn{2}{c}{Disjoint gallery (100-way)} \\
    \cmidrule(lr){2-3}\cmidrule(lr){4-5}\cmidrule(lr){6-7}\cmidrule(l){8-9}
    Method & Top-1 & Top-5 & Top-1 & Top-5 & Top-1 & Top-5 & Top-1 & Top-5 \\
    \midrule
    SAMGA (frozen)
      & \res{25.65}{34.55} & \res{56.75}{67.35} & \res{33.77}{45.13} & \res{69.10}{78.13}
      & \res{6.70}{8.40} & \res{23.38}{27.15} & \res{10.93}{13.82} & \res{34.77}{40.08} \\
    \midrule
    SATTC (SAW)~\cite{huang2026sattc}
      & \res{30.45}{31.90} & \res{61.65}{62.90} & \res{35.53}{37.83} & \res{68.90}{70.13}
      & \res{8.93}{9.80} & \res{\snd{27.78}}{28.55} & \res{\fst{13.60}}{14.37} & \res{37.88}{39.27} \\
    SCORE~\cite{cui2026score}
      & \res{41.35}{47.80} & \res{71.25}{79.85} & \res{\snd{49.57}}{57.40} & \res{81.37}{88.13}
      & \res{8.98}{10.83} & \res{26.40}{\snd{30.68}} & \res{12.98}{\fst{16.03}} & \res{\fst{38.82}}{\fst{43.75}} \\
    \midrule
    GA
      & \res{31.90}{39.88} & \res{63.42}{71.68} & \res{39.31}{49.56} & \res{73.44}{81.16}
      & \res{6.13}{7.83} & \res{21.20}{25.11} & \res{9.81}{12.89} & \res{31.00}{36.39} \\
    TENT~\cite{wang2021tent}
      & \res{28.85}{28.85} & \res{47.13}{53.43} & \res{34.21}{41.96} & \res{66.84}{75.49}
      & \res{3.94}{4.95} & \res{16.61}{21.18} & \res{8.77}{11.33} & \res{29.08}{34.15} \\
    TCR~\cite{li2025tcr}
      & \res{49.40}{49.42} & \res{66.28}{67.75} & \res{39.43}{46.69} & \res{74.39}{81.08}
      & \res{9.13}{9.26} & \res{23.93}{25.47} & \res{10.69}{12.72} & \res{32.41}{35.91} \\
    CLIPTTA~\cite{lafon2025cliptta}
      & \res{\snd{56.77}}{\snd{57.12}} & \res{\fst{83.20}}{\snd{83.22}} & \res{49.33}{\snd{59.04}} & \res{82.33}{\snd{88.86}}
      & \res{9.06}{9.43} & \res{25.42}{26.90} & \res{10.62}{12.82} & \res{32.48}{36.84} \\
    UATTA~\cite{zhang2026uatta}
      & \res{22.00}{22.45} & \res{34.08}{35.72} & \res{25.06}{30.59} & \res{56.23}{63.26}
      & \res{4.00}{5.14} & \res{14.55}{17.31} & \res{9.81}{12.42} & \res{31.90}{36.58} \\
    PASLE~\cite{hu2025selective}
      & \res{49.38}{50.23} & \res{69.57}{74.93} & \res{49.50}{58.42} & \res{\snd{82.40}}{88.13}
      & \res{\snd{9.57}}{\snd{10.90}} & \res{27.71}{30.62} & \res{12.87}{15.61} & \res{\snd{38.22}}{\snd{43.06}} \\
    \midrule
    \rowcolor{oursbg}
    \textbf{HSTTA (Ours)}
      & \res{\fst{59.22}}{\fst{61.22}} & \res{\snd{80.18}}{\fst{83.58}} & \res{\fst{53.81}}{\fst{63.44}} & \res{\fst{86.08}}{\fst{91.29}}
      & \res{\fst{10.18}}{\fst{10.96}} & \res{\fst{28.45}}{\fst{31.56}} & \res{\snd{13.13}}{\snd{15.83}} & \res{37.33}{42.38} \\
    \bottomrule
    \end{tabularx}
\end{table*}

\subsection{Iterative Re-localization with Reciprocal Hierarchical Backoff}
\label{sec:relocalization}
Adaptation changes query--gallery matches, while pseudo-label training may reinforce incorrect predictions.
We therefore periodically re-localize queries and introduce Reciprocal Hierarchical Backoff (RHB) to check fine-grained supervision through image-to-query matching (Fig.~\ref{fig:framework}(c,e)).

At the start of adaptation and every $t$ epochs, we recompute matching scores and probabilities for all queries using the current encoder and apply Climbing with the fixed threshold $\widehat\theta$.
Each gallery image selects the query with the highest CSLS score.
The indices of images selecting $q_i$ form:
\begin{equation}
    \mathcal B_i
    =
    \left\{
    j:\ i=\operatorname*{arg\,max}_{1\le h\le N^Q}s_{hj}
    \right\}.
\end{equation}
% Starting from the Climbing node $v_i$, if $\mathcal S(v_i)\cap\mathcal B_i=\varnothing$, we follow parent links until reciprocal support exists or a coarse node is reached.
% Queries already at coarse nodes remain unchanged.
% Thus, retained leaf predictions satisfy mutual top-1 matching, while mid-level nodes contain at least one image selecting the query.
% Queries backed off to coarse nodes adapt solely via distribution alignment.
% We then update $v_i$ and $\pi_i$, rebuild anchors and pseudo-labels, and optimize the EEG encoder with the objective in Sec.~\ref{sec:objective}.
Starting from the Climbing node $v_i$, RHB repeatedly replaces $v_i$ with its parent while $\mathcal S(v_i)\cap\mathcal B_i=\varnothing$ and $v_i$ is not coarse.
Retained leaf predictions satisfy mutual top-1 matching, while mid-level predictions contain an image selecting the query.
Queries at coarse nodes receive only distribution alignment.
We then update $v_i$ and $\pi_i$, rebuild anchors and pseudo-labels, and optimize the EEG encoder with the objective in Sec.~\ref{sec:objective}.

\section{Experiments}

\subsection{Experimental Setup}
\label{sec:setup}

\textbf{Datasets and evaluation protocols.}
We evaluate cross-subject retrieval on THINGS-EEG2~\cite{gifford2022large} and Alljoined-1.6M~\cite{xu2025alljoined} under leave-one-subject-out.
Both test sets contain 200 unseen concepts, each with one image and approximately 20 EEG trials per session across four sessions.
% We keep each model's pretraining, EEG preprocessing, and visual features, using official EEG preprocessing for Alljoined-1.6M.
Following SCORE~\cite{cui2026score}, we evaluate standard 200-way retrieval and transfer to an unseen gallery.
For the latter, the test gallery is randomly split three times into disjoint 100-concept subsets $A$ and $B$; we adapt using only $A$'s images and unlabeled EEG, then freeze the model for 100-way retrieval on $B$.
In this protocol, the Ward tree is built on $A$ only, with $K\in\{1,2,4,8,16,32,100\}$ and mid-level nodes at $K\in\{16,32\}$.
We report final-epoch Top-1 and Top-5 accuracy under cosine and CSLS ($k=10$) ranking, averaged over held-out subjects and three adaptation seeds; unseen-gallery results are further averaged over the three splits, giving nine runs per subject.

\textbf{Baselines and implementation details.}
Baselines include training-free methods SATTC~\cite{huang2026sattc} and SCORE~\cite{cui2026score}, and learnable TTA methods TENT~\cite{wang2021tent}, TCR~\cite{li2025tcr}, CLIPTTA~\cite{lafon2025cliptta}, UATTA~\cite{zhang2026uatta}, and PASLE~\cite{hu2025selective}.
For controlled comparisons, we run learnable baselines offline and add HSTTA's global alignment term $\mathcal L_{\mathrm{dist}}^{\mathrm{glob}}$ to their objectives.
PASLE's source-domain validation is replaced by CSCC at HSTTA's $\alpha$ for initial threshold calibration, whose reliability is analyzed in Sec.~\ref{ana:cscc}.
Following SCORE, we use SAMGA~\cite{jiang2026subject} as the main pretrained model and optimize EEG-side LoRA (rank $16$, $\alpha_{\mathrm{LoRA}}=16$) and normalization-layer affine parameters with AdamW at a constant $10^{-3}$ learning rate.
For THINGS-EEG2 and Alljoined-1.6M, respectively, we use $(\tau,\alpha,t)=(0.02,0.75,5)$ and $(0.002,0.50,5)$, gradient-norm ratios of $0.1$ and $1$ for $\gamma$, and $200$ and $50$ epochs.
Preprocessing, visual features, baseline implementations, and hyperparameter selection are detailed in the supplemental material.

\begin{table}[t]
\centering
\caption{HSTTA component ablation on THINGS-EEG2 with SAMGA.
$\mathcal{L}_{\mathrm{dist}}^{\mathrm{loc}}$ is the local distribution-alignment term in Eq.~\eqref{eq:dist_align}.}
\label{tab:ablation}
\footnotesize
\setlength{\tabcolsep}{2pt} 
\renewcommand{\arraystretch}{1.05} 
    \begin{tabularx}{\linewidth}{@{}l*{2}{>{\centering\arraybackslash}X}c@{}} 
    \toprule
    Variant & Top-1 & Top-5 & $\Delta$Top-1 \\
    \midrule
    \multicolumn{4}{@{}l}{\textit{Standard (200-way)}} \\
    \rowcolor{oursbg}
    \textbf{HSTTA (full)}                             & \res{\textbf{59.22}}{\textbf{61.22}} & \res{\textbf{80.18}}{\textbf{83.58}} & -- \\
    w/o $\mathcal{L}_{\mathrm{dist}}^{\mathrm{loc}}$  & \res{50.32}{53.50} & \res{73.58}{79.07} & \dres{-8.9}{-7.7} \\
    w/o $\mathcal{L}_{\mathrm{leaf}}$                 & \res{46.78}{50.82} & \res{73.10}{77.67} & \dres{-12.4}{-10.4} \\
    w/o $\mathcal{L}_{\mathrm{sup}}$ ($\gamma{=}0$)   & \res{37.48}{41.92} & \res{66.85}{71.97} & \dres{-21.7}{-19.3} \\
    w/o re-localization                               & \res{41.30}{46.45} & \res{69.13}{74.97} & \dres{-17.9}{-14.8} \\
    w/o RHB                                           & \res{55.83}{57.43} & \res{76.65}{80.08} & \dres{-3.4}{-3.8} \\
    \midrule
    \multicolumn{4}{@{}l}{\textit{Disjoint gallery (100-way)}} \\
    \rowcolor{oursbg}
    \textbf{HSTTA (full)}                             & \res{\textbf{53.81}}{\textbf{63.44}} & \res{\textbf{86.08}}{\textbf{91.29}} & -- \\
    w/o $\mathcal{L}_{\mathrm{dist}}^{\mathrm{loc}}$  & \res{51.67}{60.83} & \res{84.68}{89.92} & \dres{-2.1}{-2.6} \\
    w/o $\mathcal{L}_{\mathrm{leaf}}$                 & \res{44.68}{54.79} & \res{79.62}{85.98} & \dres{-9.1}{-8.7} \\
    w/o $\mathcal{L}_{\mathrm{sup}}$ ($\gamma{=}0$)   & \res{42.26}{52.48} & \res{77.36}{84.88} & \dres{-11.6}{-11.0} \\
    w/o re-localization                               & \res{48.29}{57.63} & \res{81.49}{87.41} & \dres{-5.5}{-5.8} \\
    w/o RHB                                           & \res{50.90}{60.53} & \res{84.26}{89.80} & \dres{-2.9}{-2.9} \\
    \bottomrule
    \end{tabularx}
\end{table}

\begin{figure*}[t]
  \centering
  \includegraphics[width=1\textwidth]{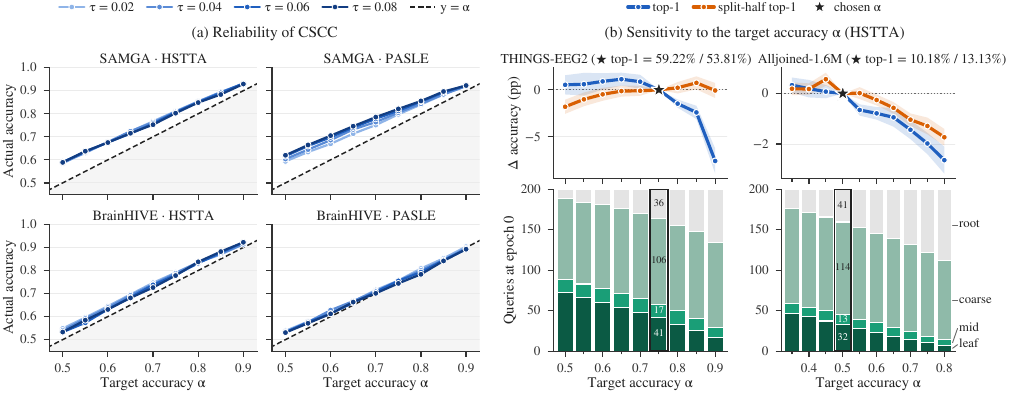}
    \caption{Reliability and sensitivity of CSCC. (a) Actual accuracy of CSCC-calibrated pseudo-labels across target $\alpha$ on THINGS-EEG2. (b) Top: HSTTA top-1 across $\alpha$, relative to the selected $\alpha$; bottom: query distribution over tree levels at epoch 0.}
  \label{fig:cscc_rel_sens}
\end{figure*}

\subsection{Main Results}
\cref{tab:main} shows that HSTTA achieves the highest Top-1 accuracy on THINGS-EEG2 under both protocols.
In standard 200-way retrieval, it improves frozen SAMGA's Top-1 accuracy from \res{25.65}{34.55} to \res{59.22}{61.22}, exceeding SCORE and the strongest learnable baseline CLIPTTA by \res{17.9}{13.4} and \res{2.5}{4.1} percentage points, respectively ($p=0.07$ ($p=0.004$) against CLIPTTA).
Although CLIPTTA achieves higher cosine Top-5 accuracy in this setting, HSTTA leads on all four disjoint-gallery metrics (all $p<0.05$), demonstrating generalization to concepts excluded from adaptation rather than merely fitting the adaptation gallery.
TENT and UATTA underperform GA under both protocols, suggesting that repeated offline updates may reinforce unreliable predictions and limit adaptation gains.

On Alljoined-1.6M, SAMGA reaches only \res{8.30}{10.57} Top-1 accuracy even on source subjects' test images.
A high target accuracy $ \alpha $ leaves most queries at coarse levels, so we lower $\alpha$ to $50\%$, yielding a level distribution comparable to THINGS-EEG2 (Sec.~\ref{ana:cscc}).
HSTTA leads on all four standard-retrieval metrics, but its gains over PASLE are not significant ($p>0.05$).
On the disjoint gallery, learnable methods generally underperform SATTC (SAW) and SCORE.
With this weak backbone, unreliable fine-grained pseudo-labels may encourage overfitting and limit transfer, favoring training-free transformations.

\begin{table}[t]
\centering
\caption{Generalization to other pretrained EEG encoders on THINGS-EEG2 under standard 200-way retrieval.}
\label{tab:backbones}
\footnotesize
\setlength{\tabcolsep}{2pt}
\renewcommand{\arraystretch}{1.05}
    \begin{tabularx}{\linewidth}{@{}cl*{2}{>{\centering\arraybackslash}X}@{}}
    \toprule
    Backbone & Method & Top-1 & Top-5 \\
    \midrule
    \multirow{5}{*}{\begin{tabular}{@{}c@{}}VE-SDN~\cite{chen2024visual}\\(arXiv 2024)\end{tabular}}
    & Frozen & \res{14.30}{18.85} & \res{39.05}{45.95} \\
    & SCORE & \res{20.10}{22.80} & \res{44.45}{50.95} \\
    & GA & \res{19.65}{22.93} & \res{46.02}{51.18} \\
    & PASLE & \res{\snd{21.17}}{\snd{23.90}} & \res{\snd{48.72}}{\snd{51.63}} \\
    & \cellcolor{oursbg}\textbf{HSTTA (Ours)} & \cellcolor{oursbg}\res{\fst{23.37}}{\fst{26.23}} & \cellcolor{oursbg}\res{\fst{50.43}}{\fst{54.43}} \\
    \midrule
    \multirow{5}{*}{\begin{tabular}{@{}c@{}}NeuroBridge~\cite{zhang2026neurobridge}\\(AAAI 2026)\end{tabular}}
    & Frozen & \res{19.40}{24.95} & \res{\snd{43.95}}{54.25} \\
    & SCORE & \res{\snd{20.95}}{\fst{27.75}} & \res{43.90}{\fst{56.80}} \\
    & GA & \res{13.72}{19.43} & \res{36.62}{46.35} \\
    & PASLE & \res{20.63}{22.42} & \res{40.95}{49.18} \\
    & \cellcolor{oursbg}\textbf{HSTTA (Ours)} & \cellcolor{oursbg}\res{\fst{22.75}}{\snd{25.58}} & \cellcolor{oursbg}\res{\fst{45.48}}{\snd{55.48}} \\
    \midrule
    \multirow{5}{*}{\begin{tabular}{@{}c@{}}BrainHIVE~\cite{ICLR2026_c5506f15}\\(ICLR 2026)\end{tabular}}
    & Frozen & \res{20.80}{27.90} & \res{44.40}{56.70} \\
    & SCORE & \res{31.05}{36.40} & \res{61.05}{69.30} \\
    & GA & \res{31.80}{36.83} & \res{63.80}{68.13} \\
    & PASLE & \res{\snd{34.80}}{\snd{38.17}} & \res{\snd{65.88}}{\snd{70.25}} \\
    & \cellcolor{oursbg}\textbf{HSTTA (Ours)} & \cellcolor{oursbg}\res{\fst{36.48}}{\fst{39.90}} & \cellcolor{oursbg}\res{\fst{67.62}}{\fst{71.02}} \\
    \bottomrule
    \end{tabularx}
\end{table}

\subsection{Ablation Study}
\label{sec:ablation}
We remove one HSTTA component at a time on THINGS-EEG2 with SAMGA (\Cref{tab:ablation}).
We report standard and disjoint-gallery Top-1 drops, respectively, under cosine (CSLS), with paired $t$-tests on cosine Top-1 over held-out subjects.
Removing $\mathcal{L}_{\mathrm{dist}}^{\mathrm{loc}}$ while keeping global alignment reduces Top-1 by \res{8.9}{7.7} and \res{2.1}{2.6} points ($p{<}0.001$; $p{=}0.07$), indicating that local distribution alignment along hierarchical paths adds gains beyond the single-image and candidate-set pseudo-label supervision used in methods such as PASLE, especially on the standard protocol.
Removing all pseudo-label supervision $\mathcal{L}_{\mathrm{sup}}$ causes the largest drops (\res{21.7}{19.3} and \res{11.6}{11.0} points), and removing only single-image pseudo-labels $\mathcal{L}_{\mathrm{leaf}}$ still costs \res{12.4}{10.4} and \res{9.1}{8.7} points (all $p{<}0.001$).
Fixing query localization after initialization reduces Top-1 by \res{17.9}{14.8} and \res{5.5}{5.8} points, and removing RHB by \res{3.4}{3.8} and \res{2.9}{2.9} points (all $p{<}0.05$), showing that both re-localization and reciprocal backoff benefit adaptation.

\subsection{Generalization across Pretrained Models}
We evaluate three recent EEG models with cross-subject retrieval capability: VE-SDN~\cite{chen2024visual}, NeuroBridge~\cite{zhang2026neurobridge}, and BrainHIVE~\cite{ICLR2026_c5506f15}, using checkpoints trained with official code and unchanged pretraining.
We calibrate $\tau$ without labels to match each frozen model's EEG--image softmax sharpness to SAMGA's, yielding $0.04$, $0.02$, and $0.05$, respectively.
All use 50 epochs and SAMGA's other hyperparameters without further tuning.
HSTTA improves all frozen backbones (\Cref{tab:backbones}), leading all four metrics on VE-SDN and BrainHIVE with respective Top-1 gains of \res{9.1}{7.4} and \res{15.7}{12.0} percentage points.
However, NeuroBridge shows smaller gains, possibly because global alignment degrades retrieval.
HSTTA still outperforms GA, suggesting hierarchical selective supervision may mitigate this degradation.

\subsection{Reliability and Sensitivity of CSCC}
\label{ana:cscc}
Per-session averaged EEG views of the same stimulus provide noisy observations of a shared evoked response, forming label-free positive pairs~\cite{shen2022contrastive}.
Inspired by using noisy observations as references~\cite{pmlr-v80-lehtinen18a}, CSCC uses cross-session consistency as a label-free proxy for hierarchical accuracy: it checks whether one view's pseudo-label set contains another's top-1 retrieval and selects the smallest $\theta$ at which this consistency reaches the target $\alpha$.
In Fig.~\ref{fig:cscc_rel_sens}(a), empirical hierarchical accuracy generally tracks $\alpha$ and meets the target in 135 of 144 THINGS-EEG2 settings, supporting the proxy's effectiveness in the evaluated settings.
However, shared prediction errors across sessions can cause consistency to overestimate hierarchical accuracy, so the target accuracy is not guaranteed.
Additional calibration analyses are provided in the supplementary material.
Fig.~\ref{fig:cscc_rel_sens}(b) examines HSTTA's sensitivity to $\alpha$.
An overly high $\alpha$ pushes queries to coarse nodes or the root (bottom), reducing the number of queries receiving $\mathcal{L}_{\mathrm{sup}}$ supervision and degrading performance (top, blue).
An overly low $\alpha$ yields finer but less accurate pseudo-labels and can hurt held-out generalization (orange, THINGS-EEG2).
Despite this accuracy--specificity trade-off~\cite{deng2012hedging}, HSTTA remains stable around the selected $\alpha$.

\begin{figure}[t]
  \centering
  \includegraphics[width=\linewidth]{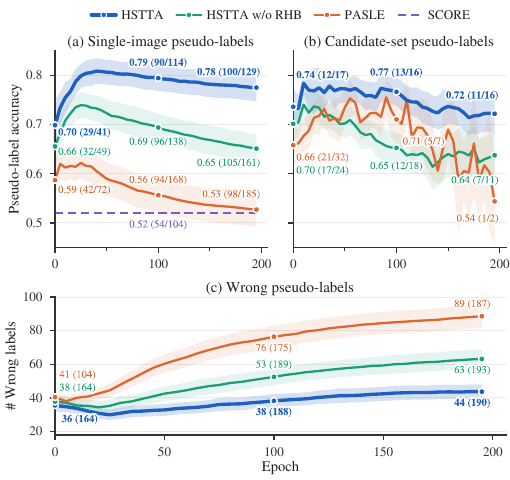}
  \caption{
    Accuracy of (a)~single-image and (b)~candidate-set pseudo-labels, annotated as acc.\ (correct/total);
    (c)~wrong pseudo-labels, annotated as wrong (supervised). Shading: s.e.\ over folds.}
  \label{fig:pl_reliability}
\end{figure}

\begin{figure}[t]
  \centering
  \includegraphics[width=\linewidth]{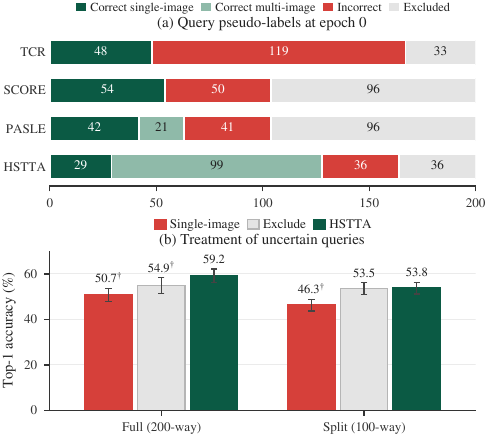}
    \caption{
      (a) Multi-image: a mid-level or coarse node; Excluded: no query-specific pseudo-label (the root for HSTTA).
      (b) Uncertain queries are those at non-leaf nodes. $^\dagger$: $p<0.05$ vs.\ HSTTA.
    }
  \label{fig:granularity}
\end{figure}

\subsection{Pseudo-Label Reliability During Adaptation}
With $\widehat\theta$ fixed by CSCC ($\alpha=0.75$), we track pseudo-labels at each re-localization, using ground truth only for evaluation.
In Fig.~\ref{fig:pl_reliability}(a), HSTTA assigns more single-image pseudo-labels ($41\!\rightarrow\!129$) while staying accurate, whereas PASLE's also grow ($72\!\rightarrow\!185$) yet lose accuracy.
HSTTA w/o RHB remains more accurate than PASLE, suggesting that hierarchical localization helps filter errors.
Its later decline supports RHB's role in limiting error accumulation.
Candidate sets show similar trends (Fig.~\ref{fig:pl_reliability}(b)).
In Fig.~\ref{fig:pl_reliability}(c), HSTTA consistently yields the fewest incorrect pseudo-labels despite supervising no fewer queries than PASLE, helping explain its performance advantage.

\subsection{Effect of Supervision Granularity}
Fig.~\ref{fig:granularity}(a) shows that HSTTA excludes only 36 queries at epoch 0 and assigns correct multi-image supervision to 99 others, whereas PASLE and SCORE each exclude 96.
Fig.~\ref{fig:granularity}(b) shows that, in standard retrieval, replacing the supervision of uncertain queries with single-image supervision or excluding them reduces Top-1 accuracy from 59.2\% to 50.7\% and 54.9\%, respectively.
These results indicate that adapting the supervision granularity to each query's confidence within the hierarchy exploits additional coarser-grained matching information and improves performance.

\section{Conclusion}
\label{sec:conclusion}
% We presented HSTTA for label-free cross-subject EEG-to-image retrieval. 
% Instead of forcing uncertain queries into unreliable single-image matches or excluding them, HSTTA localizes each query on a gallery hierarchy to match supervision granularity to its prediction confidence, combining CSCC for unsupervised threshold calibration with RHB to mitigate error reinforcement. 
% We presented HSTTA for label-free cross-subject EEG-to-image retrieval, matching supervision granularity to each query's confidence through a gallery hierarchy, with CSCC for unsupervised threshold calibration and RHB to mitigate pseudo-label error reinforcement.

We presented HSTTA for label-free cross-subject EEG-to-image retrieval, using a gallery hierarchy to match supervision granularity to each query's prediction confidence and construct adaptation objectives accordingly.
CSCC calibrates the global confidence threshold without labels, while RHB mitigates pseudo-label error reinforcement.
Analyses show that HSTTA improves over learnable TTA such as PASLE by producing fewer incorrect pseudo-labels and preserving useful matching information from uncertain queries through coarser supervision.
Extensive experiments demonstrate the effectiveness of HSTTA, which offers a new approach to label-free cross-subject adaptation.

\section*{LLM Usage Statement}
\label{sec:llm_usage_statement}
OpenAI’s ChatGPT was used to assist with editing and refining author-written text, preparing related work and citations, implementing code, designing and analyzing experiments, and developing research ideas. All AI-assisted content incorporated into this work was reviewed and verified by the authors, who take full responsibility for the manuscript and its findings.
{
    \small
    \bibliographystyle{ieeenat_fullname}
    \bibliography{main}
}
\clearpage
\maketitlesupplementary
\setcounter{section}{0}
\setcounter{figure}{0}
\setcounter{table}{0}
\renewcommand{\thesection}{S\arabic{section}}
\renewcommand{\thefigure}{S\arabic{figure}}
\renewcommand{\thetable}{S\arabic{table}}
\renewcommand{\theHsection}{S\arabic{section}}
\renewcommand{\theHfigure}{S\arabic{figure}}
\renewcommand{\theHtable}{S\arabic{table}}

\section{Experimental Details}
\label{sec:additional_exp_detail}

\subsection{Datasets and Preprocessing}
\paragraph{THINGS-EEG2.}
THINGS-EEG2~\cite{gifford2022large} contains EEG recordings from $10$ subjects viewing natural object images in a rapid serial visual presentation paradigm.
Signals were acquired at $1{,}000$~Hz with a $64$-channel EASYCAP arranged according to the 10--10 system, online band-pass filtered at $0.1$--$100$~Hz, and referenced online to Fz, yielding $63$ recorded channels.
The training set contains $1{,}654$ concepts with $10$ images per concept, each presented $4$ times, totaling $16{,}540$ training images per subject.
The test set contains $200$ concepts disjoint from the training set, each represented by a single image presented $80$ times, $20$ in each of $4$ sessions.
Following the standard preprocessing of~\cite{song2024decoding}, epochs are extracted from $-200$ to $1{,}000$~ms relative to stimulus onset, baseline-corrected using the $200$~ms pre-stimulus interval, cropped to $0$--$1{,}000$~ms, downsampled to $250$~Hz, and whitened with multivariate noise normalization.
We keep the model-specific EEG preprocessing and image feature extraction of each pretrained model unchanged from its original implementation.
For TTA and evaluation, the $80$ repetitions of each test stimulus are averaged into one query, yielding $200$ queries per subject.
CSCC is applied once before adaptation to calibrate $\widehat{\theta}$ from the four $20$-trial session averages.

\paragraph{Alljoined-1.6M.}
Alljoined-1.6M~\cite{xu2025alljoined} contains EEG recordings from $20$ subjects viewing the same THINGS images as THINGS-EEG2 in a rapid serial visual presentation paradigm across $4$ sessions.
Signals were acquired at $256$~Hz with a consumer-grade $32$-channel Emotiv Flex~2 system.
Each subject views the $16{,}540$ training images about $4$ times each and the $200$ test images about $80$ times each, roughly $20$ per session.
Following the official preprocessing pipeline~\cite{xu2025alljoined}, epochs from $-200$ to $1{,}000$~ms are baseline-corrected, downsampled to $250$~Hz, whitened with session-wise multivariate noise normalization, and cropped to $0$--$1{,}000$~ms.
Trials with missing triggers are discarded, and for any subject whose released data deviate from this pipeline, we regenerate the data from the raw recordings using the official code.
Queries and CSCC session averages are formed as for THINGS-EEG2.

\begin{figure*}[t]
  \centering
  \includegraphics[width=\linewidth]{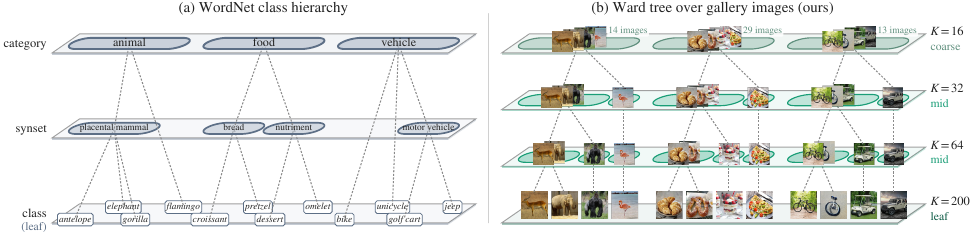}
  \caption{Comparison of a WordNet class hierarchy and a Ward tree.
    (a)~WordNet hierarchy of the selected test concepts.
    (b)~Ward tree of one held-out subject on THINGS-EEG2, built from SAMGA image embeddings.}
  \label{fig:wardtree}
\end{figure*}

\begin{figure}[t]
  \centering
  \includegraphics[width=\linewidth]{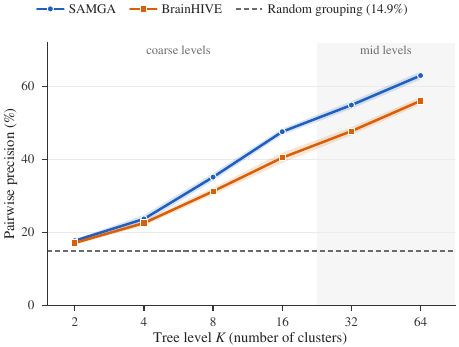}
    \caption{Pairwise precision at each level of the Ward tree on THINGS-EEG2.
  Mean $\pm$ standard error over 10 folds.}
  \label{fig:wardtreepp}
\end{figure}

\subsection{Baseline Implementations}
All methods start from the same frozen pretrained model and use the same queries, gallery, and evaluation under cosine and CSLS~\cite{lample2018word} ranking with $k = 10$.
Under the disjoint-gallery protocol, alignment statistics, fitted transformations, and model adaptation use only the EEG and images of subset $A$, while subset $B$ is used only for $100$-way evaluation.
Training-free baselines keep the encoder parameters fixed and transform only the frozen embeddings:
\begin{itemize}
    \item \textbf{SATTC}~\cite{huang2026sattc} recalibrates retrieval scores through whitening, density-adaptive CSLS, and mutual-rank structural priors.
    Following SCORE~\cite{cui2026score}, we use only its subject-adaptive whitening (SAW) step, which whitens target EEG embeddings with their own mean and shrunk covariance, as the full method with default hyperparameters underperforms frozen SAMGA in our setting.
    \item \textbf{SCORE}~\cite{cui2026score} comprises Recovery-Aware Source Training and Coordinate Recovery at Deployment.
    As all methods share the same frozen pretrained model, we use only the latter, which applies moment matching as in Eq.~(3) and a regularized orthogonal Procrustes rotation fitted on mutual nearest-neighbor pairs under CSLS.
\end{itemize}
Learnable baselines construct training objectives from label-free query--gallery predictions and update the encoder parameters offline:
\begin{itemize}
    \item \textbf{Global Alignment (GA)} adapts the encoder with $\mathcal{L}_{\mathrm{dist}}^{\mathrm{glob}}$ alone, aligning the per-dimension mean and standard deviation of rescaled EEG embeddings with those of the gallery without pseudo-labels.
    All other learnable baselines add $\mathcal{L}_{\mathrm{dist}}^{\mathrm{glob}}$ to their own losses, with $\gamma$ set as in Eq.~(15).
    \item \textbf{TENT}~\cite{wang2021tent} minimizes the entropy of each query's predictions over the full gallery.
    \item \textbf{TCR}~\cite{li2025tcr} treats each query and its top-1 image as a pseudo-pair, builds a reference queue from the least-shifted pairs early in adaptation, and uses this queue to jointly optimize a weighted entropy, a uniformity regularizer on query embeddings, and a term matching the modality mean gap to the queue's estimate.
    \item \textbf{CLIPTTA}~\cite{lafon2025cliptta} computes a soft contrastive entropy within each batch using each query's top-1 image as its pseudo-label, while maximizing the marginal entropy over the gallery to prevent prediction collapse.
    We remove its memory bank.
    \item \textbf{UATTA}~\cite{zhang2026uatta} uses the frozen model once to select queries whose forward and backward retrievals agree, estimate their uncertainty from the discrepancy between bidirectional retrieval probabilities, and take each query's top-1 image and several hard negatives as candidates.
    It then minimizes the candidate entropy weighted by inverse uncertainty.
    \item \textbf{PASLE}~\cite{hu2025selective} calibrates its probability threshold with CSCC at HSTTA's $\alpha$, replacing the source-domain validation of the original method.
    With the same scoring as HSTTA, queries whose top-1 probability exceeds the second-highest by at least the threshold receive single-image pseudo-labels; the remaining queries take images within the threshold of the top-1 probability as candidate sets under the CC loss~\cite{feng2020provably}; and queries whose candidate sets cover the entire gallery are excluded from supervision.
    The partition is updated every $t$ epochs as the threshold decreases linearly by $0.1$ over training, and the memory bank is removed.
\end{itemize}

\subsection{Implementation Details and Hyperparameter Selection}
\paragraph{Environment.} 
All test-time adaptation methods and evaluations are implemented in Python~3.13 and PyTorch~2.7.1 with CUDA~11.8, and the complete list of dependencies and versions will be provided in our released repository.
Each experiment is run on a single NVIDIA RTX A6000 GPU.

\paragraph{SAMGA Pretraining and Visual Features.}
SAMGA~\cite{jiang2026subject} takes multi-layer visual features as inputs to its image branch.
Following Shallow Alignment~\cite{du2026deep}, on which SAMGA builds, we extract the mean patch-token features from layers $20$, $24$, $28$, $32$, and $36$ of \texttt{InternViT-6B-448px-V2.5}~\cite{chen2024internvl}.
At test time, these five layers are fused using only the global layer weights learned during training, following the official cross-subject setting.
On both benchmarks, we pretrain SAMGA under the leave-one-subject-out protocol with its official cross-subject configuration and use the final checkpoint after the full $50$ epochs.
EEG and image embeddings are retrieved in a $512$-dimensional shared space.

\paragraph{Temperature Selection.}
The temperature $\tau$ is shared by the matching probabilities in Eq.~(5), the training logits in Eqs.~(12)--(13), and the predictive probabilities of the learnable baselines.
On THINGS-EEG2, we set $\tau$ without labels according to the sharpness of the frozen model's matching probabilities.
For each held-out subject, we compute the matching probabilities $\mathbf{p}_i$ between the $200$ queries and $200$ gallery images via Eqs.~(3)--(5), and take $\exp(H(\mathbf{p}_i))$ as the effective number of candidate images for query $q_i$, which reflects how many images the model is effectively choosing among.
This quantity equals $1$ when the probability mass concentrates on a single image and $k$ when it is uniform over $k$ images, and it increases monotonically with $\tau$.
We therefore find, for each held-out subject, the $\tau$ at which the median effective number over all queries equals $4$, and take the median across held-out subjects.
This rule yields $0.0216$ for SAMGA, which we round to $\tau = 0.02$, and after the same rounding yields $0.04$, $0.02$, and $0.05$ for VE-SDN, NeuroBridge, and BrainHIVE in Sec.~4.4 of the main paper.

\paragraph{Loss Weight $\gamma$.}
The weight $\gamma$ in Eq.~(15) is computed once at epoch $0$.
Using the initialized model and query localization, we compute the $\ell_2$ norms of the gradients of $\mathcal{L}_{\mathrm{dist}}$ and $\mathcal{L}_{\mathrm{sup}}$ with respect to all trainable parameters, and set $\gamma = r\,\|\nabla\mathcal{L}_{\mathrm{dist}}\|_2 / \|\nabla\mathcal{L}_{\mathrm{sup}}\|_2$, where $r$ is the preset gradient-norm ratio.
This value remains fixed throughout adaptation.
At the start of adaptation, the gradient norm of $\gamma\mathcal{L}_{\mathrm{sup}}$ is thus $r$ times that of $\mathcal{L}_{\mathrm{dist}}$, making $\gamma$ independent of the absolute scales of the two losses.
The learnable baselines follow the same rule, computing $\gamma$ from $\mathcal{L}_{\mathrm{dist}}^{\mathrm{glob}}$ and their own losses.

\section{Semantic Structure of the Hierarchical Tree}
We provide additional implementation details for constructing the gallery hierarchical tree described in Sec.~3.2.
Taking SAMGA as an example, we apply Euclidean Ward clustering to the raw 512-dimensional embeddings from the frozen image branch.
SAMGA trains a separate image projection head on each fold's source subjects, so we build one Ward tree per leave-one-subject-out fold from its projected embeddings.
Using embeddings directly from a fixed pretrained image backbone would allow all folds to share one tree per gallery.
Each tree is built solely from image embeddings and remains fixed during adaptation.
\cref{fig:wardtree} contrasts two types of hierarchies. 
\cref{fig:wardtree}(a) shows a WordNet-based~\cite{miller1995wordnet} class hierarchy for selected concepts, the type of hierarchy used by DARTS~\cite{deng2012hedging} and HSC~\cite{goren2024hierarchical} on ImageNet. 
\cref{fig:wardtree}(b) shows the actual Ward tree of one held-out subject on THINGS-EEG2, from which we select four test images each of animals, food, and vehicles to illustrate how they merge level by level from single images at $K{=}200$ into three nodes at $K{=}16$. 
Each circle marks the node containing the enclosed images.

To quantify whether the Ward tree places semantically similar images in the same node, we use the 53 high-level categories of THINGSplus~\cite{stoinski2024thingsplus} as a reference and compute the pairwise precision~\cite{christopher2008introduction} at each level $K$, \ie, the fraction of within-node image pairs that share at least one high-level category.
Randomly grouping the 200 test images gives an expected value of 14.9\%.
As shown in \cref{fig:wardtreepp}, the Ward trees built from the image-branch embeddings of SAMGA~\cite{jiang2026subject} and BrainHIVE~\cite{ICLR2026_c5506f15} exceed chance at all levels, with pairwise precision increasing toward finer levels and reaching 63.0\% and 56.0\% at $K{=}64$, respectively.
Thus, the Ward tree groups semantically similar images without category labels, supporting the use of tree nodes as semantic neighborhoods for EEG queries.

\begin{algorithm}[t]
\small
\caption{Cross-Session Consistency Calibration}
\label{alg:cscc}
\begin{algorithmic}[1]
\Require Queries $\{q_i\}$ from $A$ sessions, gallery embeddings $\{z^G_j\}$,
  tree $\mathcal{H}$, encoder $f_{\mathrm{eeg}}$, temperature $\tau$, grid $\Theta$, target accuracy $\alpha$
\Ensure Calibrated threshold $\hat{\theta}$
\For{$a = 1, \dots, A$}
  \State $\bar{x}^a_i \gets$ average of the $R^a_i$ trials of $q_i$ in session $a$
  \State $z^{Q,a}_i \gets f_{\mathrm{eeg}}(\bar{x}^a_i)$
  \State $p^a_{ij} \gets$ Eqs.~(3)--(5) on $\{z^{Q,a}_i\}_i$ and $\{z^G_j\}_j$
  \State $P^a_i(v) \gets \sum_{j \in S(v)} p^a_{ij}$
  \State $j^a_i \gets \arg\max_j p^a_{ij}$
\EndFor
\For{$\theta \in \Theta$ in ascending order}
  \State $v^a_i(\theta) \gets$ \Call{Climbing}{$P^a_i, \theta$}
  \State $\hat{C}(\theta) \gets \frac{1}{N^Q A(A-1)} \sum_{i,\, a \neq b}
         \mathbbm{1}\big[ j^b_i \in S(v^a_i(\theta)) \big]$
  \If{$\hat{C}(\theta) \geq \alpha$}
    \State \Return $\hat{\theta} \gets \theta$
  \EndIf
\EndFor
\Statex
\Function{Climbing}{$P_i, \theta$}
  \State $v \gets \arg\max_{v' \in \mathcal{Y}} P_i(v')$
  \While{$P_i(v) < \theta$}
    \State $v \gets \mathrm{parent}(v)$
  \EndWhile
  \State \Return $v$
\EndFunction
\end{algorithmic}
\end{algorithm}

\begin{figure}[t]
  \centering
  \includegraphics[width=\linewidth]{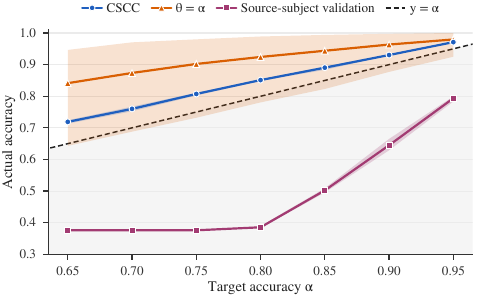}
  \caption{Threshold selection on THINGS-EEG2.
  Lines show the 10-fold mean accuracy averaged over $\tau \in \{0.02, 0.04, 0.06, 0.08\}$, with shading for the range across $\tau$.}
  \label{fig:cscc_threshold}
\end{figure}

\begin{figure}[ht]
  \centering
  \includegraphics[width=\linewidth]{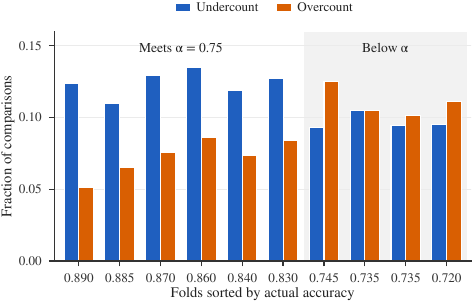}
  \caption{CSCC undercounts and overcounts per fold for SAMGA at $\alpha = 0.75$.}
  \label{fig:cscc_errorbar}
\end{figure}

\section{CSCC Algorithm and Analysis}
\subsection{CSCC Algorithm}
\cref{alg:cscc} summarizes CSCC, which uses only the target subject's unlabeled EEG, the fixed gallery embeddings, and the Ward tree.
The first loop computes matching probabilities and top-1 predictions for all queries within each session.
The second loop scans $\Theta$ in ascending order, computes $\hat{C}(\theta)$ from the per-session Climbing nodes, and returns the first $\theta$ with $\hat{C}(\theta) \geq \alpha$.
Steps indexed by $i$ and $a$ are applied to all queries and sessions.
We use a uniform grid over $[0,1]$ with a step of $0.001$ and finer steps near $0$ and $1$.
As $\theta$ increases, Climbing stops at a higher node on the same path, so $\hat{C}(\theta)$ is nondecreasing in $\theta$, and $\hat{C}(1) = 1$ since all queries reach the root.
The scan thus always returns $\hat{\theta}$ in Eq.~(9).

\begin{table*}[ht]
\centering
\caption{Cross-subject retrieval accuracy of the non-learned moment matching (MM) of SCORE~\cite{cui2026score}, \ie, Eq.~(3) of the main paper, on eight pretrained models. Accuracy is reported as cosine (CSLS) ranking in \%. $\Delta$ denotes the difference between MM and the frozen model.}
\label{tab:supp_mm}
\small
\setlength{\tabcolsep}{3pt}
\setlength{\aboverulesep}{0pt}
\setlength{\belowrulesep}{0pt}
\renewcommand{\arraystretch}{1.05}
\begin{tabularx}{\textwidth}{@{}l
  *{2}{>{\hsize=.85\hsize\centering\arraybackslash}X} >{\hsize=1.3\hsize\centering\arraybackslash}X
  *{2}{>{\hsize=.85\hsize\centering\arraybackslash}X} >{\hsize=1.3\hsize\centering\arraybackslash}X@{}}
\toprule
 & \multicolumn{3}{c}{Top-1} & \multicolumn{3}{c}{Top-5} \\
\cmidrule(lr){2-4} \cmidrule(l){5-7}
Pretrained model & Frozen & MM & $\Delta$ & Frozen & MM & $\Delta$ \\
\midrule
NICE~\cite{song2024decoding}              & \res{8.60}{10.75}  & \res{8.35}{10.05}  & \dres{-0.25}{-0.70} & \res{25.90}{30.55} & \res{22.60}{27.90} & \dres{-3.30}{-2.65}   \\
ATMS~\cite{NEURIPS2024_ba5f1233}          & \res{8.30}{12.05}  & \res{5.70}{10.40}  & \dres{-2.60}{-1.65} & \res{25.05}{32.85} & \res{16.60}{28.50} & \dres{-8.45}{-4.35}   \\
UBP~\cite{wu2025bridging}                 & \res{12.65}{19.35} & \res{6.75}{10.75}  & \dres{-5.90}{-8.60} & \res{33.20}{44.85} & \res{18.95}{29.35} & \dres{-14.25}{-15.50} \\
NeuroBridge~\cite{zhang2026neurobridge}   & \res{19.40}{24.95} & \res{17.70}{25.40} & \dres{-1.70}{+0.45} & \res{43.95}{54.25} & \res{42.40}{54.05} & \dres{-1.55}{-0.20}   \\
\midrule
VE-SDN~\cite{chen2024visual}              & \res{14.30}{18.85} & \res{19.20}{23.15} & \dres{+4.90}{+4.30} & \res{39.05}{45.95} & \res{45.15}{50.80} & \dres{+6.10}{+4.85}   \\
BrainHIVE~\cite{ICLR2026_c5506f15}        & \res{20.80}{27.90} & \res{29.85}{35.45} & \dres{+9.05}{+7.55} & \res{44.40}{56.70} & \res{60.40}{65.60} & \dres{+16.00}{+8.90}  \\
Shallow Alignment~\cite{du2026deep}       & \res{21.65}{33.00} & \res{27.60}{38.55} & \dres{+5.95}{+5.55} & \res{50.70}{65.10} & \res{57.80}{71.35} & \dres{+7.10}{+6.25}   \\
SAMGA~\cite{jiang2026subject}             & \res{25.65}{34.55} & \res{29.25}{37.60} & \dres{+3.60}{+3.05} & \res{56.75}{67.35} & \res{62.70}{71.25} & \dres{+5.95}{+3.90}   \\
\bottomrule
\end{tabularx}
\end{table*}

\begin{table*}[t]
\centering
\caption{Cross-subject retrieval accuracy (\%) of TTA methods on eight pretrained models under cosine (CSLS) ranking. MM$\downarrow$ and MM$\uparrow$ denote that MM lowers and raises the frozen accuracy, respectively. }
\label{tab:supp_tta}
\small
\setlength{\tabcolsep}{4pt}
\setlength{\aboverulesep}{0pt}
\setlength{\belowrulesep}{0pt}
\renewcommand{\arraystretch}{1.05}
\begin{tabularx}{\textwidth}{@{}c l *{4}{>{\centering\arraybackslash}X} >{\columncolor{oursbg}[\tabcolsep][0pt]\centering\arraybackslash}X@{}}
\toprule
 & Pretrained model & Frozen & SCORE & GA & PASLE & HSTTA \\
\midrule
\multicolumn{6}{@{}l}{\textit{(a) Top-1}} & \\
\multirow{4}{*}{\rotatebox[origin=c]{90}{\scriptsize MM$\downarrow$}}
 & NICE~\cite{song2024decoding}              & \res{\textbf{8.60}}{\textbf{10.75}}  & \res{7.40}{8.65}   & \res{1.68}{1.95}   & \res{2.12}{2.53}   & \res{1.98}{2.45}   \\
 & ATMS~\cite{NEURIPS2024_ba5f1233}          & \res{\textbf{8.30}}{\textbf{12.05}}  & \res{3.75}{5.00}   & \res{7.77}{10.65}  & \res{7.38}{8.85}   & \res{7.83}{10.15}  \\
 & UBP~\cite{wu2025bridging}                 & \res{\textbf{12.65}}{\textbf{19.35}} & \res{6.40}{7.70}   & \res{4.07}{7.07}   & \res{3.62}{6.15}   & \res{5.53}{8.33}   \\
 & NeuroBridge~\cite{zhang2026neurobridge}   & \res{19.40}{24.95} & \res{20.95}{\textbf{27.75}} & \res{13.72}{19.43} & \res{20.63}{22.42} & \res{\textbf{22.75}}{25.58} \\
\cmidrule(l){2-7}
\multirow{4}{*}{\rotatebox[origin=c]{90}{\scriptsize MM$\uparrow$}}
 & VE-SDN~\cite{chen2024visual}              & \res{14.30}{18.85} & \res{20.10}{22.80} & \res{19.65}{22.93} & \res{21.17}{23.90} & \res{\textbf{23.37}}{\textbf{26.23}} \\
 & BrainHIVE~\cite{ICLR2026_c5506f15}        & \res{20.80}{27.90} & \res{31.05}{36.40} & \res{31.80}{36.83} & \res{34.80}{38.17} & \res{\textbf{36.48}}{\textbf{39.90}} \\
 & Shallow Alignment~\cite{du2026deep}       & \res{21.65}{33.00} & \res{33.60}{\textbf{42.50}} & \res{22.10}{32.68} & \res{33.42}{34.22} & \res{\textbf{36.30}}{38.48} \\
 & SAMGA~\cite{jiang2026subject}             & \res{25.65}{34.55} & \res{41.35}{47.80} & \res{31.90}{39.88} & \res{49.38}{50.23} & \res{\textbf{59.22}}{\textbf{61.22}} \\
\midrule
\multicolumn{6}{@{}l}{\textit{(b) Top-5}} & \\
\multirow{4}{*}{\rotatebox[origin=c]{90}{\scriptsize MM$\downarrow$}}
 & NICE~\cite{song2024decoding}              & \res{\textbf{25.90}}{\textbf{30.55}} & \res{17.50}{22.95} & \res{6.15}{6.57}   & \res{8.02}{8.98}   & \res{7.32}{8.35}   \\
 & ATMS~\cite{NEURIPS2024_ba5f1233}          & \res{\textbf{25.05}}{\textbf{32.85}} & \res{12.40}{18.90} & \res{22.52}{29.73} & \res{21.62}{28.05} & \res{22.50}{27.97} \\
 & UBP~\cite{wu2025bridging}                 & \res{\textbf{33.20}}{\textbf{44.85}} & \res{17.30}{23.30} & \res{13.05}{21.93} & \res{13.35}{19.83} & \res{14.98}{23.12} \\
 & NeuroBridge~\cite{zhang2026neurobridge}   & \res{43.95}{54.25} & \res{43.90}{\textbf{56.80}} & \res{36.62}{46.35} & \res{40.95}{49.18} & \res{\textbf{45.48}}{55.48} \\
\cmidrule(l){2-7}
\multirow{4}{*}{\rotatebox[origin=c]{90}{\scriptsize MM$\uparrow$}}
 & VE-SDN~\cite{chen2024visual}              & \res{39.05}{45.95} & \res{44.45}{50.95} & \res{46.02}{51.18} & \res{48.72}{51.63} & \res{\textbf{50.43}}{\textbf{54.43}} \\
 & BrainHIVE~\cite{ICLR2026_c5506f15}        & \res{44.40}{56.70} & \res{61.05}{69.30} & \res{63.80}{68.13} & \res{65.88}{70.25} & \res{\textbf{67.62}}{\textbf{71.02}} \\
 & Shallow Alignment~\cite{du2026deep}       & \res{50.70}{65.10} & \res{\textbf{61.00}}{\textbf{74.70}} & \res{51.57}{65.57} & \res{53.15}{60.68} & \res{58.00}{64.05} \\
 & SAMGA~\cite{jiang2026subject}             & \res{56.75}{67.35} & \res{71.25}{79.85} & \res{63.42}{71.68} & \res{69.57}{74.93} & \res{\textbf{80.18}}{\textbf{83.58}} \\
\bottomrule
\end{tabularx}
\end{table*}

\begin{figure*}[t]
  \centering
  \includegraphics[width=\textwidth]{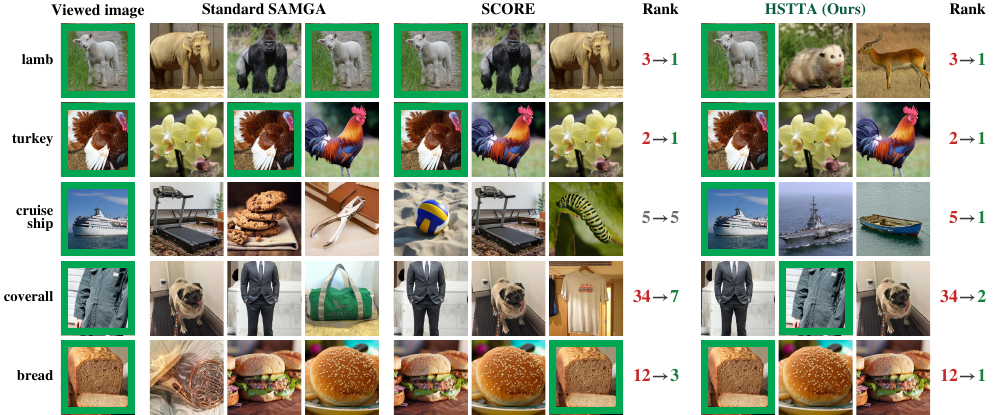}
  \caption{Top-3 retrieved images on held-out subject~10 of THINGS-EEG2.
  Green borders mark the viewed image, and Rank shows its rank among the 200 gallery images before and after adaptation.}
  \label{fig:supp_retrieval_examples}
\end{figure*}

\begin{figure}[t]
  \centering
  \includegraphics[width=\linewidth]{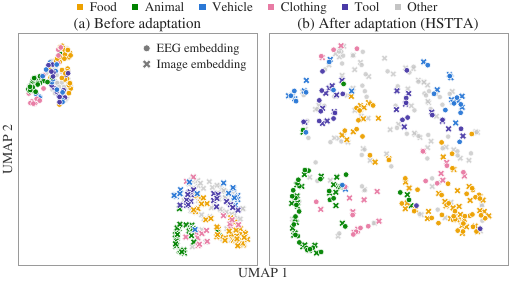}
  \caption{UMAP visualization of $\ell_2$-normalized EEG and image embeddings for all 200 test images of held-out subject~10 before (a) and after (b) HSTTA.}
  \label{fig:umap}
\end{figure}

\subsection{Calibration Analysis}
We compare CSCC with two unsupervised alternatives:
(a) $\theta = \alpha$, which treats the node confidence $P_i(v)$ in Eq.~(6) as calibrated and uses the target accuracy as the threshold;
(b) source-subject validation, which averages the hierarchical accuracy curves of the nine source subjects computed with ground-truth labels and selects the first $\theta$ reaching $\alpha$.
\cref{fig:cscc_threshold} shows the results of HSTTA with SAMGA before adaptation.
The 10-fold mean accuracy of CSCC exceeds $\alpha$ in all settings and varies by at most $0.014$ across $\tau$.
In contrast, $\theta = \alpha$ varies by up to $0.30$ across $\tau$, falling below the target at $\tau = 0.02$ and exceeding it at $\tau \geq 0.04$.
Source-subject validation falls below the target in all settings: as the source subjects are seen during pretraining, their top-1 accuracy under CSLS ranking reaches $0.807$, versus $0.376$ on the target subject, so for $\alpha \leq 0.75$ it selects $\theta = 0$ and reduces to top-1 prediction.
Thus, source-calibrated thresholds do not transfer under subject shift, whereas CSCC, using only the target subject's unlabeled EEG, meets the target stably across $\tau$ in the evaluated settings.

Moreover, CSCC uses cross-session consistency as a proxy for accuracy, which introduces two opposite biases.
An undercount occurs when $S(v^a_i)$ contains the ground-truth image but $j^b_i \notin S(v^a_i)$, which lowers $\hat{C}$; an overcount occurs when $S(v^a_i)$ excludes the ground-truth image but $j^b_i \in S(v^a_i)$, which raises $\hat{C}$.
Shared errors across sessions increase overcounts.
When the top-1 prediction of one session is wrong, the other session retrieves the same wrong image with probability $0.159$, compared with $0.005$ by chance.
\cref{fig:cscc_errorbar} shows the undercounts and overcounts of the ten folds for SAMGA at $\alpha = 0.75$.
In the six folds that meet the target, undercounts exceed overcounts and CSCC is conservative.
In the four folds below the target, overcounts are at least as frequent as undercounts, so $\hat{C}$ overestimates the accuracy, which drops to $0.720$--$0.745$.
Hence, CSCC is a conservative label-free proxy overall, but it cannot guarantee the target accuracy for every subject when shared errors dominate.

\section{Generalization across Pretrained Models}
We extend the experiments in Sec.~4.4 of the main paper to more representative recent EEG visual decoding methods on THINGS-EEG2 to examine the performance of TTA methods across them. 
Specifically, we evaluate eight pretrained models, namely NICE~\cite{song2024decoding}, ATMS~\cite{NEURIPS2024_ba5f1233}, VE-SDN~\cite{chen2024visual}, UBP~\cite{wu2025bridging}, NeuroBridge~\cite{zhang2026neurobridge}, BrainHIVE~\cite{ICLR2026_c5506f15}, Shallow Alignment~\cite{du2026deep}, and SAMGA~\cite{jiang2026subject}. 
We observe that these models respond inconsistently to TTA. 
We first conduct a basic analysis with the non-learned moment matching (MM) of SCORE~\cite{cui2026score}, \ie, Eq.~(3) of the main paper, which aligns EEG embeddings to image embeddings dimension-wise.
As shown in \cref{tab:supp_mm}, MM does not improve, and even reduces, cross-subject accuracy on NICE, ATMS, UBP, and NeuroBridge, whereas it improves accuracy on VE-SDN, BrainHIVE, Shallow Alignment, and SAMGA.
We conjecture that the modality gap between the EEG and image embeddings of NICE and similar models lies not only in global location and scale but also in the relative relations among samples. 
Since MM only shifts and rescales the queries, it leaves these relations unchanged and may thus fail to improve, or even degrade, retrieval.

Moreover, we conduct full TTA experiments on all eight models. 
All implementation details of HSTTA follow Sec.~4.4 of the main paper, and we report the accuracy at the last epoch.
Results are shown in \cref{tab:supp_tta}. 
Since HSTTA also builds on MM, it yields clear gains on all pretrained models where MM is effective, consistently outperforming PASLE and outperforming SCORE on all of them except Shallow Alignment.
For NeuroBridge, on which MM is ineffective, HSTTA remains effective, suggesting that hierarchical selective supervision may mitigate the degradation caused by global alignment. 
On NICE, ATMS, and UBP, SCORE, PASLE, and HSTTA all degrade performance. 
We leave improvements for such models to future work.

\section{Qualitative Results}
\subsection{Image Retrieval Examples}
Fig.~\ref{fig:supp_retrieval_examples} shows retrieval examples before and after TTA on THINGS-EEG2 with SAMGA as the pretrained model. 
The examples are from held-out subject~10, and all three methods rank the 200 gallery images by cosine similarity.
Before adaptation, SAMGA ranks none of the viewed images first. 
SCORE moves two of them to the top, whereas HSTTA ranks four first and raises the remaining \textit{coverall} from 34th to 2nd, behind \textit{suit}, another clothing item. 
In these examples, most of the other images that HSTTA ranks near the top are also semantically related to the viewed image. 
For \textit{cruise ship}, HSTTA ranks \textit{aircraft carrier} and \textit{boat} right after it, whereas SAMGA and SCORE rank unrelated images such as \textit{treadmill} near the top. 
Together with the disjoint-gallery results in Tab.~1 of the main paper, this indicates that HSTTA improves how the encoder maps the new subject's EEG into the semantic image space, rather than merely fitting pseudo-labels on the adaptation set. 
Moreover, the gains of HSTTA are not limited to queries with correct single-image pseudo-labels.
\textit{Turkey} and \textit{coverall} never receive a correct single-image pseudo-label during adaptation and are supervised mainly at coarse nodes, yet they still reach ranks 1 and 2, respectively. 
This is consistent with the findings on supervision granularity in Sec.~4.7 of the main paper.

\subsection{EEG and Image Embedding Visualization}
Taking held-out subject~10 as an example, we visualize the $\ell_2$-normalized embeddings of all 200 test images and their corresponding EEG responses with UMAP (Fig.~\ref{fig:umap}).
Before adaptation, the EEG embeddings form a separate cluster, clearly apart from the image embeddings.
After HSTTA, the EEG embeddings move into the region occupied by the image embeddings, and most of them lie close to images of the same category.
This indicates that HSTTA narrows the gap between new subject's EEG and image representations.


\end{document}